%% file: acl_arxiv.tex
\documentclass[11pt]{article}

\usepackage[preprint]{acl}
\usepackage{tcolorbox}
\usepackage{listings}

\usepackage{times}
\usepackage{graphicx}
\usepackage{multirow}
\usepackage{textcomp}
\usepackage{stfloats}
\usepackage{url}
\usepackage{verbatim}
\usepackage{amsmath, amsfonts}
\usepackage{booktabs}
\usepackage[T1]{fontenc}
\usepackage{pifont}
\usepackage{algorithm}
\usepackage{algorithmic}
\usepackage{booktabs}  
\usepackage{enumitem}
\usepackage{xcolor}
\usepackage{amssymb} 

\usepackage{xcolor}
\usepackage{amssymb} 

\usepackage{array}
\usepackage[table,xcdraw]{xcolor}
\usepackage{graphicx} 
\usepackage{booktabs} 
\usepackage{amssymb}

\usepackage[utf8]{inputenc}
\usepackage{booktabs}
\usepackage{graphicx}
\usepackage{amssymb}
\usepackage{titling}

\definecolor{rblue1}{HTML}{E1F5FE} 
\definecolor{rblue2}{HTML}{B3E5FC} 
\definecolor{rblue3}{HTML}{81D4FA} 
\definecolor{rblue4}{HTML}{4FC3F7} 
\definecolor{bgA}{HTML}{FFFDE7} 
\definecolor{bgB}{HTML}{FFF59D} 
\definecolor{bgC}{HTML}{FFD54F} 
\definecolor{bgD}{HTML}{FFB74D} 
\definecolor{mysuppcolor}{RGB}{197,116,23}
\definecolor{suppcolor}{RGB}{0,128,128} 
\definecolor{lowblue}{RGB}{230,240,255}
\definecolor{midblue}{RGB}{180,205,240}
\definecolor{highblue}{RGB}{120,165,220}
\newcommand{\supp}{\textit{supplementary material}} 

\usepackage[utf8]{inputenc}
\newcommand{\benchmarkname}{FinMTM}
\usepackage{microtype}

\usepackage{inconsolata}

\usepackage{graphicx}

\usepackage{calc}  
\usepackage{booktabs}  
\usepackage{pifont}    

\newcommand{\spadesym}{\textsuperscript{\ding{171}}}



\title{
\includegraphics[height=0.9cm]{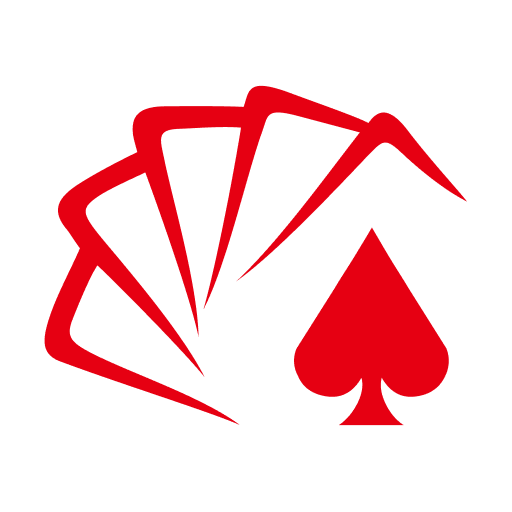}
\hspace{-0.2cm}
FinMTM: A Multi-Turn Multimodal Benchmark for \\ Financial Reasoning and Agent Evaluation
}

\author{
  \textbf{Chenxi Zhang}$^{1,2}$\thanks{Equal contribution.}
  \quad
  \textbf{Ziliang Gan}$^{1,3}$\footnotemark[1]
  \quad
  \textbf{Liyun Zhu}$^{1}$\footnotemark[1]
  \quad
  \textbf{Youwei Pang}$^{4}$
  \\
  \textbf{Qing Zhang}$^{5}$
  \quad
  \textbf{Rongjunchen Zhang}$^{1}$\spadesym
  \\
  $^{1}$HiThink Research \quad
  $^{2}$Wuhan University \quad
  $^{3}$Zhejiang University
   \quad
  \\
  $^{4}$Nanyang Technological University \quad
  $^{5}$Shanghai Institute of Technology
  \\
  \textbf{Correspondence:} \texttt{zhangrongjunchen@myhexin.com}
}

\definecolor{rowhighlight}{HTML}{FFF5EE} 
\definecolor{checkcolor}{HTML}{C0392B}   
\definecolor{dashcolor}{HTML}{666666}

\newcommand{\yes}{\textcolor{checkcolor}{\ding{51}}} 
\newcommand{\no}{\textcolor{dashcolor}{\ding{55}}}

\begin{document}
\maketitle

\begingroup
  \renewcommand\thefootnote{\ding{171}}%
  \footnotetext[3]{Corresponding author.}%
\endgroup

\begin{abstract}
The financial domain poses substantial challenges for vision-language models (VLMs) due to specialized chart formats and knowledge-intensive reasoning requirements. 
However, existing financial benchmarks are largely single-turn and rely on a narrow set of question formats, limiting comprehensive evaluation in realistic application scenarios.
To address this gap, we propose \benchmarkname, a multi-turn multimodal benchmark that expands diversity along both data and task dimensions. 
On the data side, we curate and annotate 11{,}133 bilingual (Chinese and English) financial QA pairs grounded in financial visuals, including candlestick charts, statistical plots, and report figures.
On the task side, \benchmarkname~ covers single- and multiple-choice questions, multi-turn open-ended dialogues, and agent-based tasks.
We further design task-specific evaluation protocols, including a set-overlap scoring rule for multiple-choice questions, a weighted combination of turn-level and session-level scores for multi-turn dialogues, and a composite metric that integrates planning quality with final outcomes for agent tasks.
Extensive experimental evaluation of 22 VLMs reveal their limitations in fine-grained visual perception, long-context reasoning, and complex agent workflows. Code is available at \url{https://github.com/HiThink-Research/FinMTM}.
\end{abstract}

\section{Introduction}

The rapid advancement of large language models (LLMs) and vision-language models (VLMs) has profoundly impacted various industries~\cite{brown2020language,openai2023gpt4,huang2023language}. In particular, the financial domain, characterized by an abundance of specialized charts, presents an ideal scenario for the application of VLMs. Complex and professional financial charts provide rigorous tests of models’ abilities in chart understanding and fine-grained visual perception. Furthermore, sophisticated tasks such as financial analysis~\cite{zhu2025findeepresearch}, quantitative factor mining~\cite{li2025rdagent}, stock prediction~\cite{lin2025retuning}, and investment decision-making require models to integrate domain-specific knowledge with reasoning capabilities, retrieving and synthesizing key information from vast datasets to yield professional insights. Additionally, agent-based tasks demand that models perform in-depth analysis and autonomously invoke tools to gather information and derive final conclusions~\cite{wu2023bloomberggpt,li2024qwen}. Collectively, these tasks pose significant challenges to current VLMs.

\begin{figure}[!t]
    \centering
    \includegraphics[width=\linewidth]{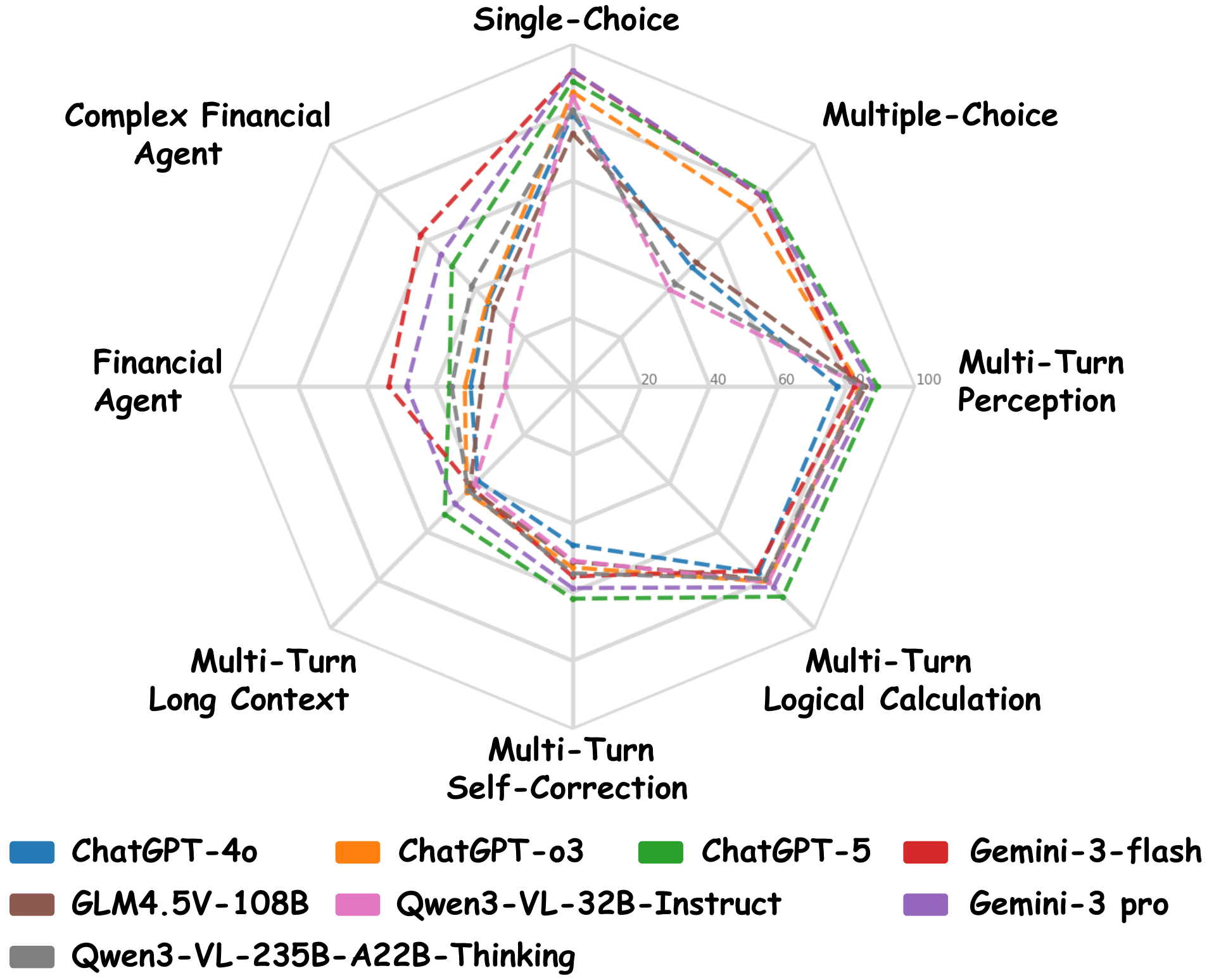}
    \caption{The radar chart of the overall performance across different dimensions.}
    \label{fig:radar}
    \vspace{-16pt}
\end{figure}

Recently, researchers have introduced some financial benchmarks, such as MME-Finance~\cite{gan2025mme}, FinMME~\cite{luo2025finmme}, FinTMMBench~\cite{zhu2025fintmmbench}, FinRAGBench-V~\cite{zhao2025finragbenchv}, and VisFinEva~\cite{liu2025visfineval}. 
Although these benchmarks evaluate the perception and reasoning capabilities of VLMs to a certain extent, they still suffer from the following three main limitations.

First, most benchmarks primarily assess VLMs through single-turn settings, relying on binary-choice questions or short free-form answers. This format tends to oversimplify complex financial reasoning and fails to fully capture the models' analytical depth.
Second, current frameworks lack a systematic evaluation of comprehensive capabilities, particularly in areas such as multi-turn interaction, multi-source information fusion, long-context reasoning, and knowledge-intensive financial analysis.
Third, existing benchmarks largely neglect financial-agent behaviors, such as iterative tool usage, strategic path planning, and result synthesis, which are critical components of real-world financial workflows.
Together, these limitations raise a key question: Can VLMs that excel in static single-turn benchmarks effectively sustain their performance in complex, interactive multi-turn scenarios and financial agent tasks involving long-horizon planning and model context protocol (MCP) interactions?

\begin{figure*}[t]
\centering
\includegraphics[width=0.99\textwidth]{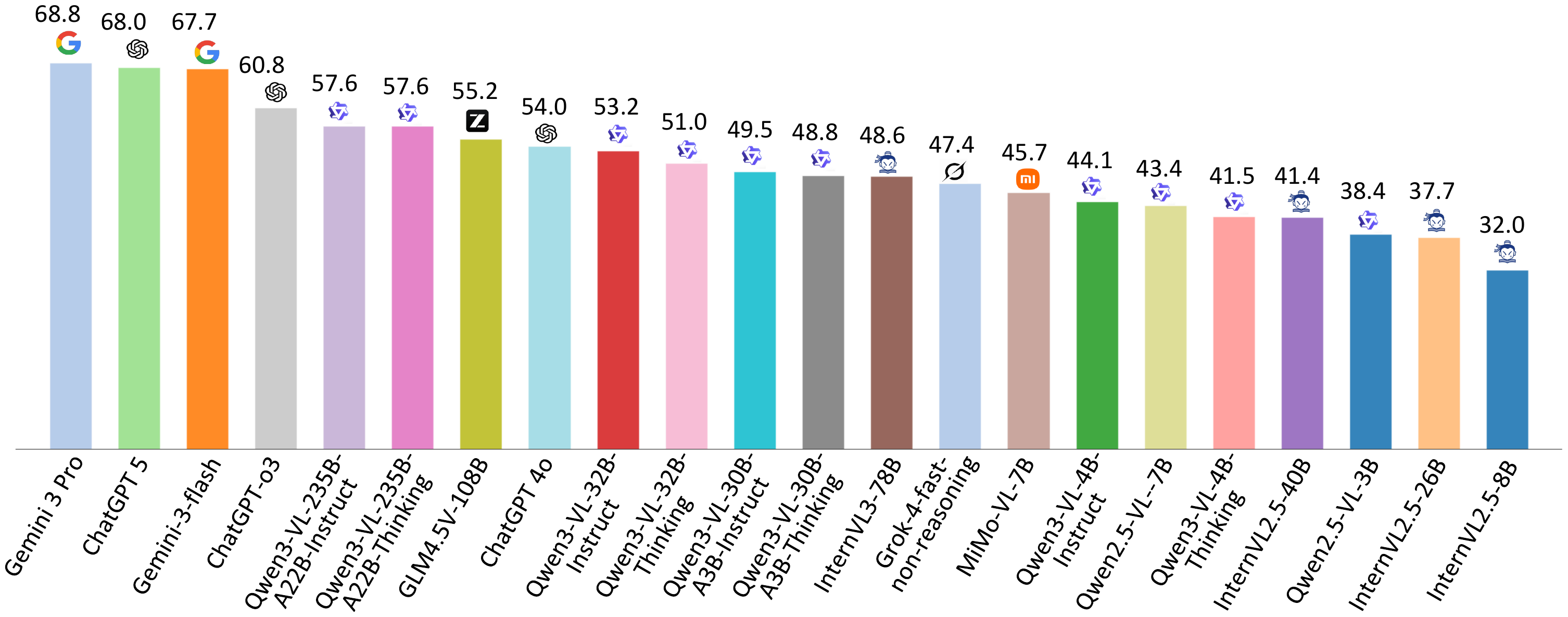}
\caption{Comparison of leading VLMs on general financial capabilities on FinMTM benchmark. The final score is calculated as the average of three task scores: objective questions, open-ended questions, and financial agent.}
\label{fig:fig1}
\end{figure*}

To bridge these gaps, we introduce \textbf{FinMTM}, a comprehensive financial multimodal benchmark designed to holistically evaluate perception, reasoning, and agentic capabilities in realistic settings. 
As illustrated in Fig.~\ref{fig:radar} and Fig.~\ref{fig:fig1}, we conduct extensive comparative experiments across a diverse set of leading VLMs on FinMTM.
The results reveal clear performance gaps across different scenarios.
Our main contributions are summarized as follows:

\begin{itemize}[leftmargin=*]
\item \textbf{A Large-Scale Bilingual Financial Benchmark.}
We construct a bilingual (Chinese and English) benchmark with over 10{,}000 samples, grounded in specialized financial visuals (\textit{e.g.}, candlestick charts, statistical diagrams, and report figures) and spanning single-/multiple-choice QA, multi-turn open-ended dialogues, and financial agent-based tasks.

\item \textbf{A Multi-Dimensional Evaluation Framework.}
We design task-specific metrics and evaluation pipelines, including strict set-based scoring for choice questions and a dual-rule LLM-judge protocol (turn-level capability scoring and session-level checklists) for open-ended and agent tasks.

\item \textbf{Extensive Evaluation and Insights.}
We evaluate 16 open-source and 6 proprietary VLMs on \benchmarkname~ and identify persistent bottlenecks in fine-grained visual grounding, complex financial reasoning, long-context consistency, and tool-augmented agent planning.
\end{itemize}

\begin{table*}[t]
\centering
\setlength{\tabcolsep}{3.5pt}
\begin{tabular}{l c ccccc ccc}
\toprule
\multirow{2}{*}{\textbf{Benchmark}} & \multirow{2}{*}{\textbf{Vol.}} & \multicolumn{5}{c}{\textbf{Task Types}} & \multicolumn{3}{c}{\textbf{Advanced Capabilities}} \\
\cmidrule(lr){3-7} \cmidrule(lr){8-10}
 & & \textbf{ST} & \textbf{SC} & \textbf{MC} & \textbf{MO} & \textbf{MLC} & \textbf{Mem} & \textbf{MS} & \textbf{FT} \\
\midrule
CFBenchmark~\cite{li2025cfbenchmark}            & 9k   & \yes & \yes & \no  & \no  & \no  & \no  & \no  & \no  \\
FinCriticalED~\cite{he2025fincriticaled}       & 0.5k & \no  & \no  & \no  & \no  & \no  & \no  & \no  & \no  \\
FinMME~\cite{luo2025finmme}                    & 11k  & \yes & \yes & \yes & \no  & \no  & \no  & \no  & \no  \\
FinMR~\cite{deng2025finmr}                     & 3.2k & \yes & \no  & \no  & \no  & \no  & \no  & \no  & \no  \\
MME-Finance~\cite{gan2025mme}                  & 2k   & \yes & \no  & \no  & \no  & \no  & \no  & \no  & \no  \\
FinRAGBench~\cite{zhao2025finragbenchv}        & 13k  & \yes & \no  & \no  & \no  & \no  & \no  & \no  & \no  \\
VisFinEval~\cite{liu2025visfineval}            & 15k  & \yes & \yes & \yes & \no  & \no  & \no  & \no  & \no  \\
\midrule
\rowcolor{rowhighlight}
\textbf{FinMTM (ours)} & \textbf{11k} & \textbf{\yes} & \textbf{\yes} & \textbf{\yes} & \textbf{\yes} & \textbf{\yes} & \textbf{\yes} & \textbf{\yes} & \textbf{\yes} \\
\bottomrule
\end{tabular}
\caption{Comparison of our FinMTM against existing multimodal financial benchmarks.
Symbols: \yes~supported; \no~not supported.
MC: multiple-choice; MO: multi-step open-ended questions;
MLC: multi-turn long-context QA;
Mem: memory capability; MS: multi-step calculation; FT: financial tool-use.}
\label{tab:comparison}
\end{table*}

\section{Related Work}
\subsection{Financial Multimodal Benchmarks}

To systematically evaluate the visual-linguistic capabilities of next-generation VLMs within the financial domain, a series of financial multimodal benchmarks have recently emerged. The overall evolutionary trajectory has proceeded from single-turn chart perception toward knowledge-intensive reasoning.
Early representative works, such as MME-Finance~\cite{gan2025mme}, FinMME~\cite{luo2025finmme}, and CFBenchmark-MM~\cite{li2025cfbenchmark}, primarily construct single-turn QA tasks based on charts and report screenshots. 
These benchmarks focus on hierarchical evaluations of chart interpretation, basic numerical reasoning, and preliminary ``perception-reasoning-cognition'' capabilities. 
However, the interaction paradigm in these early works is largely confined to single-turn QA, failing to encompass scenarios involving long documents, multi-page citations, or tool-chain integration.
Subsequently, FinMR~\cite{deng2025finmr} shifted the focus toward knowledge-intensive high-order reasoning. 
By introducing structured reasoning chains into complex valuation and risk analysis tasks, it reinforces the characterization of professional domain knowledge and multi-step reasoning. Furthermore, FinRAGBench-V~\cite{zhu2025fintmmbench} incorporates multi-page PDF documents—along with their embedded tables and charts—into the evaluation framework, explicitly measuring the factual consistency of multimodal RAG with key numerical and temporal data. 
Concurrently, VisFinEval~\cite{liu2025visfineval} organizes approximately 15k samples based on front-, middle-, and back-office business workflows, emphasizing scenario-aware understanding and holistic decision support.

As shown in Tab.~\ref{tab:comparison}, existing financial VLM benchmarks focus primarily on single-turn tasks, offering limited insight into multi-turn reliability or long-context reasoning. Essential agentic capabilities, such as iterative tool use and strategic planning, also remain under-evaluated. To bridge these gaps, we propose FinMTM, a comprehensive framework that evaluates objective QA, multi-turn dialogues, and tool-augmented agent tasks.

\section{FinMTM Benchmark}

\subsection{Task Definition}

We organize the required financial capabilities into three representative evaluation tasks, namely objective questions, open-ended questions, and financial agent. 
Specifically, objective questions include single-choice (SC) and multiple-choice (MC) formats. Open-ended questions are designed in a multi-turn setting and further categorized into four subtasks: comprehension, calculation, self-correction, and memory. For the financial agent task, we offer a suite of MCP tools to support tool calling and planning processes.
Together, these tasks form a progressive evaluation framework, spanning from basic perception, through in-depth analytical understanding, to interactive decision-making and agentic behavior. All tasks are provided in both Chinese and English versions.

\begin{figure*}[!t]
    \centering
    \includegraphics[width=\linewidth]{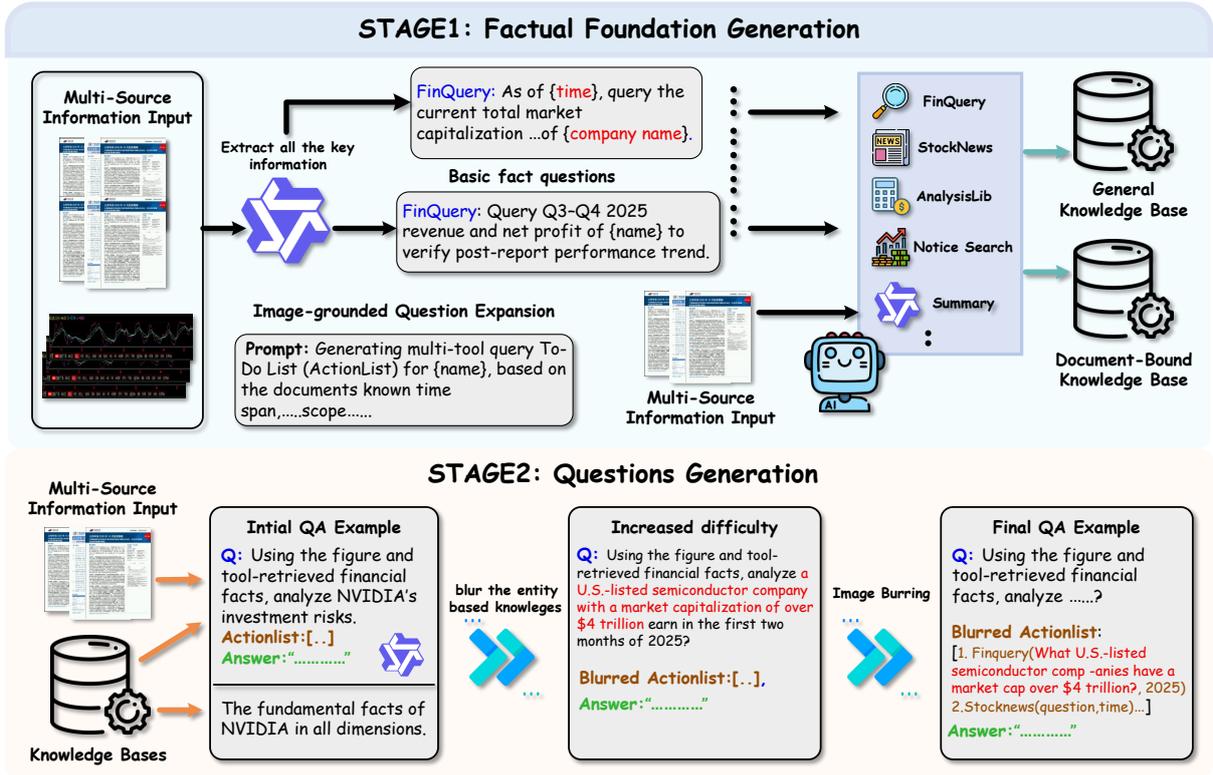}
    \caption{The proposed data synthesis pipeline tailored for financial agent question generation.
}
    \label{fig:agentic}
\end{figure*}

\subsection{Dataset Construction}

\noindent \textbf{Data Curation.} To construct FinMTM, we collect a diverse set of financial images and documents from real-world industry workflows, drawing data from four key channels: macroeconomic and industry research reports, individual stock research reports, publicly available company financial data, and financial media images. Leveraging authentic information across U.S. and Chinese stock markets, our dataset enables robust, realistic benchmarking.

\noindent\textbf{Objective Question Generation.}
We first curate a high-quality seed set of 1982 single-turn QA pairs, with every sample rigorously validated by domain experts and defined as as ground truth. Next, we use Gemini 3 Pro~\cite{google2025gemini3prosystemcard} to generate plausible distractors for each seed question, from which we build a set of standard single-choice (SC) questions. For the construction of multiple-choice (MC) question, we frame the task as a negative selection paradigm: instead of choosing the correct answer, the model is instructed to select multiple incorrect options. The design of MC question effectively reduces the probability of guessing correctly, thereby presenting a significantly more rigorous challenge to the models.



\noindent\textbf{Open-ended Questions Generation.}
We developed a pipeline utilizing Gemini 3 Pro to construct expert-level multi-turn interactions based on a principle of progressive complexity.
Initially, we employ an automated process with unified prompting rules to standardize the structure, difficulty, and evaluation objectives across diverse open-ended subtasks. As detailed in Tab.~\ref{tab:statistics}, these subtasks are categorized into four progressive levels ($L_1$- $L_4$), spanning cognitive demands from basic perception to long-term memory. 
{Notably, this taxonomy is defined at the session level}, characterizing the overall cognitive demand of an entire multi-turn conversation rather than that of any individual turn.
However, rule-based generation often lacks the nuance of real-world human-computer interaction. To address this, we integrate manual refinement to naturalize the dialogues. By introducing colloquialisms and realistic ambiguities while maintaining structural control, we ensured the interactions closely mimic authentic financial workflows.

\noindent\textbf{Financial Agentic Data Generation.}
To construct an agent evauation dataset containing reference tool calls and ground truth answers, we designed a two-stage data synthesis pipeline, as illustrated in Fig.~\ref{fig:agentic}.
First, we construct two knowledge bases for financial agentic data generation: a general knowledge base $K_1$ and a document-related knowledge base $K_2$.
These knowledge bases store verifiable financial facts and tool-call parameters.
$K_1$ covers fundamental financial facts across companies and time ranges, and supports retrieval via company entity indices. 
$K_2$ contains financial factual knowledge related to document content.
Second, we generate initial Q\&A pairs based on the images and the knowledge base.
Subsequently, to further increase task difficulty, we apply a \textbf{fuzz} strategy that systematically obfuscates entity priors. 
Specifically, we substitute specific entities within the questions using the knowledge base. Finally, we inspect the visual content and mask regions that leak critical information, such as company names.
Further implementation details are provided in Section~\ref{sec:Knowledge Base Construction Details} of \supp{}.

\subsection{Dataset Statistics}

Our benchmark consists of three main tasks with 11,133 questions with 3600 images and 400 PDFs. It covers stocks from both the U.S. and Chinese A-share markets, spanning multiple industries, and reflects diverse real-world financial settings.
The objective questions include 1,982 single-choice and 1,982 multiple-choice questions, assessing a model’s basic understanding of financial multimodal information under clearly defined constraints. Open-ended questions contain 6,169 samples, with an average of 4.25 interaction rounds per sample and 4,963 average input tokens.
Finally, the financial agent task includes 1,000 samples, all requiring external financial tools. About 73\% of the tasks can be completed using a single tool across multiple calls, while the remaining 27\% require multiple tools. This setting reflects realistic financial analysis and decision-making workflows.
Detailed statistics are provided in Tab.~\ref{tab:statistics} and supplementary material.

\begin{table}[ht]
\centering
\setlength{\tabcolsep}{1pt}
\begin{tabular}{lc}
\toprule
\textbf{Statistic} & \textbf{Number} \\
\midrule
\textbf{Objective Questions} & 3964 \\
\hspace{1em}- Single Choice & 1982 \\
\hspace{1em}- Multiple Choice& 1982 \\
\midrule
\textbf{Open-Ended Questions} & 6169 \\
\hspace{1em}\textbf{Comprehension ($L_1$).} & 2082 \\
\hspace{2em}- Entity Recognition &  \\
\hspace{2em}- Spatial Awareness &  \\
\addlinespace 
\hspace{1em}\textbf{Calculation ($L_2$).} & 1893 \\
\hspace{2em}- Multi-step Numerical Calculation &  \\
\hspace{2em}- Chart numerical Estimation & \\
\addlinespace
\hspace{1em}\textbf{Self-correction ($L_3$).} & 1210 \\
\hspace{2em}- Adversarial Robustness &  \\
\hspace{2em}- Logical Consistency &  \\
\addlinespace
\hspace{1em}\textbf{Memory ($L_4$).} & 984 \\
\hspace{2em}- Cross-page Entity Linking &  \\
\hspace{2em}- Long Context &  \\
\hspace{2em}- Multi-source Knowledge Fusion &  \\
\midrule
\textbf{Financial Agent} & 1000 \\
\hspace{1em}- Single Agent & 728 \\
\hspace{1em}- Multiple Agent & 272  \\
\midrule
\bottomrule
\end{tabular}
\caption{Statistical breakdown of FinMTM across distinct task categories and corresponding subtasks.}
\label{tab:statistics}
\end{table}

\begin{table*}[t]
\centering
\renewcommand{\arraystretch}{1.15}
\setlength{\tabcolsep}{4.2pt} 
\resizebox{\linewidth}{!}{%
\setlength{\aboverulesep}{0pt}
\setlength{\belowrulesep}{0pt}
\input{table/compare}
}

\caption{Performance of VLMs on FinMTM. Models marked with * are evaluated on a subset of FinMTM due to provider policy constraints. \textbf{Com.}, \textbf{Cal.}, \textbf{SelfCorr.}, \textbf{Mem.} denote open-ended tasks of \textit{comprehension}, \textit{calculation}, \textit{self-correction}, and \textit{memory consistency}, respectively. \textbf{w fuzz} and \textbf{w/o fuzz} indicate financial agent tasks with/without ambiguous input conditions. Top two results per subtask are bolded and underlined, respectively.}
\label{tab:main_results}
\end{table*}

\subsection{Comprehensive Financial Evaluation}
We present tailored evaluation frameworks for three distinct tasks: objective question, open-ended question, and financial agent. The robustness of our framework is validated via human consistency experiment, detailed experimental analysis is provided in Section~\ref{sec:evaluator_analysis} of the \supp{}.


\subsubsection{Objective Questions Evaluation}

We adopt an set-overlap scoring rule for single and multiple choice questions.
Each question $i$ is associated with a ground-truth answer set $G_i \subseteq \mathcal{O}$
and a model-predicted answer set $P_i \subseteq \mathcal{O}$, where $\mathcal{O}$ denotes
the set of all candidate options. The score for question $i$ is defined as
\begin{equation}
\mathrm{Score}_i =
\begin{cases}
0, & P_i \setminus G_i \neq \varnothing, \\[6pt]
\dfrac{|P_i \cap G_i|}{|G_i|}, & \text{otherwise}.
\end{cases}
\end{equation}

This formulation enforces a strict no-overselection rule: any incorrect selection
results in a zero score; otherwise, partial credit is assigned proportionally.
Single-choice questions correspond to the special case $|G_i|=1$.

\subsubsection{Open-Ended Questions Evaluation}
We propose a dual-rule evaluation strategy to assess model performance at both the turn and session levels.
Detailed prompts for the LLM-as-a-judge at both levels are provided in \supp{}. For a multi-turn dialogue
$\mathcal{D} = \{r_1, r_2, \dots, r_T\}$, the evaluation is conducted as follows.
\paragraph{Turn-level Evaluation.}
Each response $r_t$ is evaluated along five fundamental financial capabilities,
visual precision (VP), financial logic (FL), data accuracy (DA),
cross-modal verification (CMV), and temporal awareness (TA).
This yields a per-turn capability score vector:
\begin{equation}
\mathbf{s}_t =
\bigl(
s_t^{\mathrm{VP}},\;
s_t^{\mathrm{FL}},\;
s_t^{\mathrm{DA}},\;
s_t^{\mathrm{CMV}},\;
s_t^{\mathrm{TA}}
\bigr),
\end{equation}
where each component $s_t^{c} \in [0,10]$ reflects the model’s performance with
respect to capability $c$.
We further aggregate the five capability scores into a turn-level score. 



\begin{equation}
S_{\mathrm{t}}(\mathcal{D})
=
\frac{1}{T}
\sum_{t=1}^{T}
\sum_{c \in \mathcal{C}}
w^{c} \, s_t^{c},
\end{equation}
where
$\mathcal{C} = \{\mathrm{VP}, \mathrm{FL}, \mathrm{DA}, \mathrm{CMV}, \mathrm{TA}\}$
represents capability dimensions, and we use uniform weights
$w^{c}=0.2$ for all $c\in\mathcal{C}$.

\paragraph{Session-level Evaluation.}
Each session is labeled with a task type $\ell \in \{\textit{Com.}, \textit{Cal.}, \textit{SelfCorr.}, \textit{Mem.}\}$ during dataset construction. 
For a given session $\mathcal{D}$, we first determine its task label $\ell$ and then compute the session-level score $S_{\mathrm{e}}(\mathcal{D})$ by scoring the whole session with the corresponding checklist $\mathcal{K}_\ell$. 
As summarized in Tab.~\ref{tab:statistics}, different dialogue types emphasize different capability dimensions.

\paragraph{Overall Score.}
To obtain a single scalar score for each dialogue, we aggregate the turn-level
and session-level evaluations in a unified manner.
The final dialogue score is obtained by a weighted combination of the two:
\begin{equation}
S_{\mathrm{final}}(\mathcal{D}) =
\alpha \, S_{\mathrm{t}}(\mathcal{D})
+
(1-\alpha)\, S_{\mathrm{e}}(\mathcal{D})
\end{equation}
where we fix $\alpha = 0.5$ to balance the turn-level and session-level evaluation.

\subsubsection{Financial Agent Evaluation}
We propose a two-stage trajectory-based evaluation framework for financial multimodal agents, assessing their multi-step, multi-dimensional interaction.

\noindent \textbf{Planning Stage.}
The model outputs a structured trajectory that specifies tool invocation paths.
Following prior agent benchmarks~\cite{qin2023toolbench,liu2024agentbench,zhou2024webarena,zhufinmcp}, 
we quantify tool usage quality with a recall-oriented $F_\beta$ score, where matching is based on functional alignment rather than tool-name string matching.
We represent each tool invocation as $\tau=\langle \text{name}, \text{args} \rangle$, where \text{args} contains core parameters such as stock tickers, date ranges, or query strings.
Let $\hat{\mathcal{T}}$ denote the predicted set and $\mathcal{T}^\star$ the reference set.
A predicted invocation $\hat{\tau}\in\hat{\mathcal{T}}$ is counted as a True Positive (TP) only if both the tool name and its core arguments are semantically consistent with an entry in $\mathcal{T}^\star$.
We compute Precision and Recall as $P=\mathrm{TP}/|\hat{\mathcal{T}}|$ and $R=\mathrm{TP}/|\mathcal{T}^\star|$, respectively, and define:
\begin{equation}
Q^{\mathrm{t}} = w_t \cdot \frac{(1+\beta^2)PR}{\beta^2P+R}, \quad \beta > 1,
\end{equation}
where $w_t=25$, $\beta=2$, $P,R \in [0,1]$, and $Q^{\mathrm{t}} \in [0,25]$.

\noindent \textbf{Summarization Stage.}
The model synthesizes the planning trajectory and tool-returned information to produce a final answer.
Both the reasoning and the final answer must be grounded in acquired visual evidence and tool outputs.
Under this evidence-constrained setting, we evaluate reasoning quality ($Q^{\mathrm{r}}\in[0,25]$) and answer correctness ($Q^{\mathrm{a}}\in[0,50]$) against the ground truth.
The final score is:
\begin{equation}
Q_{\mathrm{final}} = Q^{\mathrm{a}} + Q^{\mathrm{r}} + Q^{\mathrm{t}} .
\end{equation}

\section{Experiment}

\begin{table}[t]
\centering
\renewcommand{\arraystretch}{1.15}
\setlength{\tabcolsep}{4.2pt} 
\resizebox{\linewidth}{!}{%
\setlength{\aboverulesep}{0pt}
\setlength{\belowrulesep}{0pt}
\input{table/compare_v2}

}
\caption{
{
Performance of VLMs under multi-turn questions, $S_e$ denotes the {session-level average score}.
}
}
\label{tab:Model_Ability}
\end{table}

\begin{table}[t]
\centering

\setlength{\tabcolsep}{3pt} 
\resizebox{\linewidth}{!}{
    \begin{tabular}{lcccccc}
    \toprule
    Models & Turn 1 & Turn 2 & Turn 3 & Turn 4 & Avg. &$\beta$ \\
    \midrule
    Qwen2.5-VL-3B              & 52.9 & 58.7 & 56.0 & 68.8 & 59.1 & \cellcolor{rblue2}4.5 \\
    Qwen2.5-VL-7B              & 61.2 & 61.9 & 65.0 & 72.5 & 65.2 & \cellcolor{rblue1}3.7 \\
    Qwen3-VL-30B-A3B-Inst.     & 61.2 & 61.2 & 74.0 & 81.6 & 69.5 & \cellcolor{rblue4}7.4 \\
    Qwen3-VL-30B-A3B-Think.    & 61.8 & 53.5 & 54.7 & 58.8 & 57.2 & -0.8 \\
    Qwen3-VL-32B-Inst.         & 69.2 & 69.2 & 77.7 & 85.5 & 75.4 & \cellcolor{rblue2}5.7 \\
    Qwen3-VL-235B-A32B         & 70.5 & 71.0 & 82.5 & 88.5 & 78.1 & \cellcolor{rblue3}6.5 \\
    Qwen3-VL-235B-A22B-Think.  &68.7 & 68.8 & 80.0 & 86.2 & 75.9 & \cellcolor{rblue3}6.4 \\
    \bottomrule
    \end{tabular}

}

\caption{Per-turn scores and trend $\beta$, where $\beta$ denotes the slope of performance improvement over turns.}
\label{tab:per_turn}
\end{table}

\begin{table}[t]
\centering
\renewcommand{\arraystretch}{1.08}
\setlength{\tabcolsep}{3.8pt}
\scriptsize
\resizebox{\columnwidth}{!}{%
\begin{tabular}{l l c c}
\toprule
\textbf{Model} & \textbf{Mode} & \textbf{SC} & \textbf{MC} \\
\midrule
\multirow{3}{*}{Qwen3-VL-32B}
& Thinking-on         & 83.4 & 46.5 \\
& Thinking-off & 85.0 & 45.8 \\
& $\Delta$ (on$-$off) & -1.6 & +0.7 \\
\midrule
\multirow{3}{*}{Qwen3-VL-30B-A3B}
& Thinking-on         & 71.5 & 49.4 \\
& Thinking-off  & 73.0 & 48.6 \\
& $\Delta$ (on$-$off) & -1.5 & +0.8 \\
\bottomrule
\end{tabular}%
}
\caption{Ablation of inference-time thinking on FinMTM. SC = single-choice, MC = multiple-choice. }
\label{tab:thinking_ablation_acmc}
\end{table}

\subsection{Experimental Setup}
\noindent\textbf{Evaluated Models.}
We evaluate a total of 22 VLMs, including both proprietary and open-source systems.
Specifically, the proprietary models include ChatGPT-4o~\cite{gpt4o}, ChatGPT-o3~\cite{openai2025o3systemcard}, ChatGPT-5~\cite{openai2025gpt5systempdf}, as well as Gemini 3 Pro~\cite{google2025gemini3prosystemcard} and Gemini 3 Flash~\cite{google2025gemini3flashsystemcard}, together with Grok-4-fast-non-reasoning~\cite{oracle_xai_grok4fast}.
The open-source models consist of the InternVL series, \textit{i.e.}, InternVL2.5-{8, 26, 40}B~\cite{internvl25} and InternVL3-78B~\cite{internvl3}, as well as the Qwen family, including Qwen2.5-VL-{3, 7}B and Qwen3-VL-{4B, 30B-A3B, 32B, 235B-A22B}-{Instruct, Thinking}~\cite{qwen25vl,qwenvl3}.
And we evaluate other representative open-source VLMs, including MiMo-VL-7B~\cite{xiaomi2025mimo} and GLM4.5V-108B~\cite{hong2025glm45v}, to further enrich model diversity.


\subsection{Experimental Results} 
Based on the results presented in Tab.~\ref{tab:main_results}, proprietary models outperform open-source counterparts across most dimensions. Notably, Gemini 3 Pro excels in open-ended tasks, leading in subtasks such as comprehension (87.5), calculation (82.8), self-correction (58.8), and memory (48.5). These results highlight Gemini 3 Pro's ability to effectively handle complex reasoning and intricate dialogues. On the other hand, ChatGPT-5 stands out for its best performance in multi-choice questions (79.6), while providing the second-best results across open-ended questions. Meanwhile, Gemini 3 Flash achieves leading scores (53.6/62.6) in fuzzed/non-fuzzed financial agent tasks, showcasing outstanding agentic capability for autonomous decision-making.

The performance gap between proprietary and open-source models becomes evident in open-ended questions and financial agent task. While open-source models (\textit{e.g.}, InternVL, Qwen VL series) compete well on objective tasks (Qwen3-VL-32B-Instruct scores 84.5 on single-choice questions and GLM4.5V-108B scores 51.0 in multi-choice questions), they degrade sharply in self-correction and memory-intensive open-ended tasks. For instance, Qwen3-VL-32B-Instruct lags behind leading proprietary models like Gemini 3 Pro (self-correction: 71.5 \textit{v.s.} 58.8; memory: 44.2 \textit{v.s.} 48.5), highlighting open-source models' limitations in complex reasoning and long-horizon memory.

In financial agent tasks, open-source models fall short in comparison to proprietary ones. For example, Qwen3-VL-32B-Thinking scores 25.1 (fuzzed) and 28.6 (non-fuzzed), while Gemini 3 Flash leads with 53.6 and 62.6. This underperformance exposes open-source models' limitations in multi-step reasoning and tool-based environments, both of which are critical for financial forecasting and decision-making. More experimental results and analysis are provided in Section~\ref{sec:Further Analysis agent} of the \supp{}.





\subsection{Analysis}

\noindent\textbf{Impact of Task Complexity on Model Performance.} 
As shown in Tab.~\ref{tab:main_results}, experimental results of open-ended questions reveal a pronounced performance decay across most models as task complexity increases($L_1$-$L_4$).
Models demonstrate robust performance at the foundational levels of Com. ($L_1$) and Cal. ($L_2$).  
For instance, Qwen3-VL-32B-Instruct achieves high scores in both categories (84.3 and 80.7, respectively).
Conversely, performance diminishes significantly as cognitive demands escalate. For the task of SelfCorr. ($L_3$) and Mem. ($L_4$), a consistent downward trend is observed across all models.  Specifically, the memory-intensive task ($L_4$), characterized by multi-image referencing, dense knowledge grounding, and long-context reasoning, remain a formidable barrier for open-source models. 
Even the top-performing model, Gemini 3 Pro, achieved a score of only 48.5, while InternVL2.5-8B scored a mere 16.7. This highlights that complex reasoning within long-context scenarios remains a significant challenge for current models.

\noindent\textbf{Detailed Breakdown of Turn-level Scores.} 
Tab.~\ref{tab:Model_Ability} details turn-level performance, showing strong results in Visual Precision and Temporal Awareness but weaknesses in Financial Logic, Data Accuracy, and Cross-modal Verification. This suggests that while current models have attained a high level of proficiency in visual perception, their capabilities in complex logical reasoning require further improvement.
Tab.~\ref{tab:per_turn} presents the scores of various models across different turns, along with the calculated slope ($\beta$) representing the trend of score changes over turns.
Overall, most VLMs benefit from extended interactions, suggesting effective contextual learning.
This gain is partly attributable to later turns that encourage models to re-attend to the visual evidence and revise earlier mistakes (\textit{e.g.}, correcting misread values or inconsistent interpretations) under the accumulated dialogue context.

\noindent\textbf{Performance Gains from the ``Thinking'' Mode.}  
We observe a task-dependent trade-off between the Thinking and Instruct variants.
Thinking improves performance in higher-entropy settings (\textit{e.g.}, financial agent tasks; 235B: $38.7 \rightarrow 41.5$), but slightly degrades in low-level perceptual tasks.
A plausible explanation is that tasks requiring fast visual--semantic alignment (\textit{e.g.}, $L_1$ comprehension and single-choice questions) benefit from direct mapping, whereas longer chain-of-thought generation may introduce distracting intermediate content and blur salient visual cues.
As shown in Tab.~\ref{tab:thinking_ablation_acmc}, disabling thinking yields a slight gain on low-entropy SC, whereas enabling thinking performs better in the higher-entropy MC.

\noindent\textbf{Error Analysis.}
We conducted an error analysis on low-scoring multi-turn instances:
For $L_1$, errors typically originate from the initial misperception of key figures or entities. These errors propagate and amplify across subsequent turns as the model fails to re-verify its outputs against visual evidence.
For $L_2$, failures primarily arise from variable-value misbinding, underspecified reference frames, or unit/scale inconsistencies (\textit{e.g.}, mistaking local extrema for global ones). This often results in logically coherent reasoning steps that nonetheless lead to incorrect computations.
For $L_3$, challenges occur when later turns introduce external facts or alter prerequisites. Models frequently fail to revise prior assumptions, leading to conclusions that are inconsistent with the updated context.
For $L_4$, failures mainly stem from weak evidence grounding within multi-page documents. Models often struggle to retrieve support from specified pages, yielding incomplete or spurious citations and filling gaps with hallucinated details. 
Detailed analysis is provided in the Section~\ref{Error Analysis supp} of~\supp{}.

\noindent\textbf{Expert Verification.}
To ensure data reliability, a panel of 15 financial experts and academic researchers verified the dataset. Furthermore, we conducted human alignment experiments to validate the evaluation accuracy for open-ended questions. Details are shown in the Section~\ref{sec:evaluator_analysis} of~\supp{}.

\section{Conclusion}
We introduce \benchmarkname, a multi-turn multimodal benchmark for evaluating comprehensive financial reasoning and agentic tool-use capabilities of VLMs. 
We conduct extensive evaluations on state-of-the-art proprietary and open-source VLMs, revealing that current models struggle with long-horizon multi-turn consistency, multi-document financial reasoning, and structured financial agent planning despite strong performance on single-turn perception and numerical tasks. 
Based on the insights and failure patterns identified in this benchmark, future work will focus on developing more robust financial VLMs with improved long-context reasoning, evidence grounding, and tool-driven decision-making abilities.

\section{Limitations}
\benchmarkname\ has several limitations.
First, although we adopt careful quality control, parts of the open-ended evaluation rely on LLM-as-a-judge, which can introduce residual subjectivity and sensitivity to prompt design; we release the scoring rubrics and prompts in the \supp{}, but improving judge robustness and calibration remains an important direction for future work.
Second, the financial agent setting currently assumes a fixed set of MCP tools and predefined reference trajectories; while our metrics attempt to assess functional alignment beyond tool-name matching, real-world tool ecosystems are broader, evolve over time, and may require evaluation under non-stationary tool availability and outputs.
Third, the long-document tasks ($L_4$) impose strict citation and formatting constraints; this mirrors compliance-oriented workflows in practice, but may underestimate capability when a response is substantively correct yet fails to follow the required protocol.
Future versions of \benchmarkname\ could incorporate stronger automatic verification, such as structured evidence extraction and citation consistency checks, and expand tool coverage to better approximate open-world financial decision environments.

\bibliography{ref}

\clearpage
\appendix

\section{Experiment Details}
We evaluate proprietary and open-source models separately.
Proprietary models are accessed through commercial APIs, while smaller open-source models are deployed locally.
All local experiments are conducted on NVIDIA H200 GPUs, and we use \texttt{vLLM} for efficient deployment and inference.
For local inference, we enable tensor parallelism across multiple GPUs and set a unified decoding configuration for fair comparison: temperature $T=0$, top-$p=1.0$, maximum prompt length of 32,768 tokens, and maximum response length of 4096 tokens.
For completeness, we report the inference prompts in Fig.~\ref{box:QA-prompt}, Fig.~\ref{box:Multiturn-prompt} and Fig.~\ref{box:agent-prompt}.

\section{Evaluator Analysis}
\label{sec:evaluator_analysis}

We conduct experiments to assess the effectiveness of different evaluators under our dual-rule evaluation protocol, which combines turn-level capability scoring and session-level checklist verification. We randomly sample 120 multi-turn instances from the open-ended split and generate corresponding outputs using Qwen3-VL-30B-A3B-Instruct as the fixed response generator. Each instance is independently scored by three experienced annotators with finance-domain expertise. The final human score is aggregated by majority vote (\textit{e.g.}, a final score of 3 for scores of 2, 3, and 3) or by mean rounding (\textit{e.g.}, a final score of 2 for scores of 1, 2, and 3). If the score variance exceeds 2 points, the instance is sent to an adjudication round to determine the final score.

\begin{figure}[!t]
    \centering
    \includegraphics[width=\linewidth]{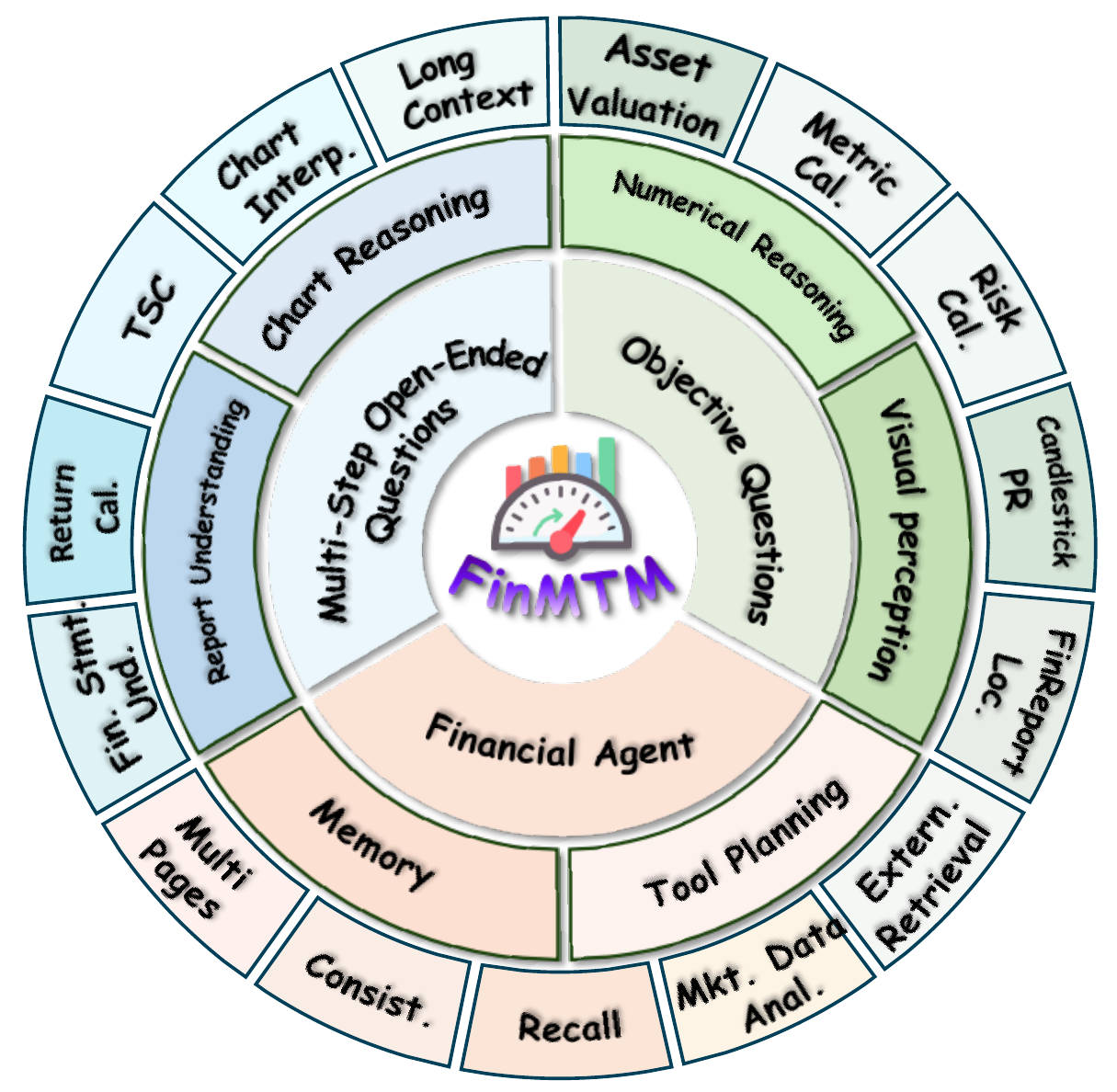}
    \caption{Overview of the proposed \benchmarkname.
    }
    \label{fig:circle}
    \vspace{-6pt}
\end{figure}

We then compute Spearman's rank correlation coefficient (Sp.) and the average absolute difference ($\Delta$) between each evaluator's scores and the human-annotated scores to quantify evaluator reliability. As shown in Tab.~\ref{tab:evaluator_agreement}, ChatGPT-4o yields the strongest alignment with human judgments, while o3 and Gemini-3-Pro also demonstrate competitive agreement. Among open-source evaluators, Qwen3-VL-235B-A22B provides comparable consistency at substantially lower evaluation cost, making it a practical alternative for large-scale benchmarking. Moreover, we observe that incorporating image inputs generally improves evaluator consistency, suggesting that direct access to visual evidence reduces ambiguity during scoring.

We further divide questions into objective and subjective categories. As shown in Tab.~\ref{tab:evaluator_cat}, the ChatGPT-4o evaluator exhibits higher agreement on objective questions (higher Sp.\ and lower $\Delta$), whereas subjective questions remain more challenging and sensitive to visual grounding and rubric interpretation. Finally, beyond the quantitative agreement study, we additionally conduct hierarchical audits by senior financial experts and academic researchers to verify that the automated evaluation is consistent with real-world financial investment research logic.

\begin{table}[t]
\centering
\caption{Agreement between evaluator scores and human-annotated scores on FinMTM. Larger Spearman's rank correlation coefficient (Sp.) and smaller average absolute difference ($\Delta$) indicate better evaluator reliability. ``Pic.'' denotes adding the original image as input during evaluation.}
\label{tab:evaluator_agreement}
\setlength{\tabcolsep}{4pt}
\begin{tabular}{lcc}
\toprule
Model & Sp. ($\uparrow$) & $\Delta$ ($\downarrow$) \\
\midrule
ChatGPT-3.5 & 0.46 & 1.42 \\
ChatGPT-4-Turbo & 0.71 & 0.95 \\
o1-preview & 0.62 & 1.12 \\
ChatGPT-4o & 0.78 (0.81) & 0.76 (0.71) \\
Gemini-3-flash (Pic.) & 0.74 (0.77) & 0.83 (0.79) \\
Gemini-3-pro (Pic.) & 0.72 (0.73) & 0.92 (0.90) \\
\bottomrule
\end{tabular}
\end{table}

\begin{table*}
\centering
\caption{Category-wise agreement of ChatGPT-4o evaluator with human scores. ``w'' and ``w/o'' denote evaluating with and without image input, respectively.}
\label{tab:evaluator_cat}
\setlength{\tabcolsep}{6pt}
\begin{tabular}{lcccccccc}
\toprule
& \multicolumn{4}{c}{Objective} & \multicolumn{4}{c}{Subjective} \\
\cmidrule(lr){2-5}\cmidrule(lr){6-9}
& Sp. & $\Delta$ & Sp. & $\Delta$ & Sp. & $\Delta$ & Sp. & $\Delta$ \\
& (w) & (w) & (w/o) & (w/o) & (w) & (w) & (w/o) & (w/o) \\
\midrule
ChatGPT-4o &
0.86 & 0.55 &
0.84 & 0.60 &
0.54 & 1.18 &
0.49 & 1.26 \\
\bottomrule
\end{tabular}
\end{table*}

\section{Further Dataset Details}
\label{sec:Further Dataset Details}

\subsection{Dataset Overview}
Fig.~\ref{fig:circle} illustrates an overview of the proposed \textbf{FinMTM} benchmark, which is designed to systematically evaluate vision language models (VLMs) under realistic financial reasoning and agentic interaction scenarios.
FinMTM organizes financial tasks into three
complementary categories, namely \emph{multi-step open-ended questions}, \emph{objective questions}, and \emph{financial agent tasks}, which covering a broad spectrum of capabilities required in real-world financial analysis. Each task category is further decomposed into fine-grained capability dimensions.
\emph{Multi-step open-ended questions} emphasize chart understanding, chart reasoning, and long-context comprehension, testing models’ ability to perform iterative reasoning over visual financial evidence.
\emph{Objective questions} focus on numerical reasoning and visual perception, including metric calculation, risk assessment, asset valuation, and candlestick pattern recognition, which require precise grounding in both visual and numerical signals.
\emph{Financial agent tasks} evaluate higher-level agentic abilities such as memory consistency, recall, tool planning, external retrieval, and multi-page reasoning, reflecting realistic multi-round financial workflows.

\subsection{Dataset Annotation}
To reduce manual annotation cost and improve efficiency, we introduce a machine-assisted annotation scheme. 
First, financial domain experts design automatic criteria to assess question quality and rationality. 
We then employ three strong vision-language models (VLMs), namely Gemini 3 Pro, ChatGPT-5, and QwenVL-235B-A22B, 
to conduct an initial evaluation of the rationality and answerability of objective and multi-step open-ended questions. 
For a given question, if all three models agree that it meets the quality criteria, it is directly accepted without further fine-grained human annotation. 
Only questions with disagreement  are forwarded to human annotators for review.

During the human annotation stage, we recruit a team of 15 annotators, including 10 junior annotators and 5 experts. 
Junior annotators with basic financial knowledge are responsible for question review, rewriting, and independent problem solving. 
The expert group consists of (i) four academic researchers from computer science and finance-related fields, all holding at least a master’s degree, 
and (ii) four finance industry professionals. 
Experts are responsible for question selection, quality assessment, and final answer verification. 
Overall, the annotation and review process requires approximately 600 hours of human effort, 
significantly reducing annotation cost while maintaining high data quality.

\subsection{Knowledge Base Construction Details}
\label{sec:Knowledge Base Construction Details}

This framework consists of three steps: knowledge base construction, initial trajectory synthesis, and entity blurring and trajectory evolution based on fact injection.

(i) We build a verifiable knowledge base supported by industrial-grade financial tools, which includes a general knowledge base $K_1$ and a document-bound knowledge base $K_2$. The knowledge base stores verifiable facts and tool call parameters. 
${K_1}$ covers basic financial facts across companies and time ranges, and supports retrieval by company entity index. 
Under the guidance of domain experts, we define common financial analysis dimensions (such as market capitalization, valuation, etc.), automatically extract company entities from financial reports and research documents, and combine "entity × dimension" into question templates to batch retrieve and accumulate corresponding verifiable facts.

For $K_2$, we focus on content expansion at the level of individual financial reports, with indexing and retrieval supported via document identifiers.
We take the explicitly stated facts in the document (such as quarterly revenue, profits, etc.) as anchor points, set structured expansion strategies and guide the model to conduct expansion reasoning along dimensions such as time and scope.
During the process, the model automatically generates tool query parameters and acquires results through the MCP interface.

(ii) By integrating multimodal information with verifiable facts obtained from $K_1$and $K_2$ based on tool calls, we explicitly constrain and guide the model to organize the facts into a coherent analysis trajectory. 
Based on this trajectory, the model undergoes reverse construction to generate corresponding financial questions and answers, obtaining an initial agent trajectory that is logically consistent and fact-verifiable.

(iii) For constructing more challenging and generalizable evaluation scenarios without compromising verifiability, we progressively incorporate tool-derived facts from the knowledge base into problem reformulation, reorganizing the original queries at both the entity and expression levels.

\begin{table}[t]
\centering
\caption{Statistics of \benchmarkname~ of multi-turn open-ended questions. Lexical diversity is measured using MTLD~\cite{McCarthyJarvis2010_MTLDVOCDHDD}.}
\label{tab:dataset_stats}
\renewcommand{\arraystretch}{0.92}
\setlength{\tabcolsep}{3pt}
\footnotesize
\begin{tabular}{p{0.68\linewidth}c}
\toprule
Statistic & Number \\
\midrule
Total conversations & 6,169 \\
Unique images & 2,427 \\
Avg./Max. turns & 4.25 / 5 \\
\midrule
Avg./Max. question length (chars) & 105.40 / 667 \\
Avg./Max. answer length (chars) & 75.65 / 1,146 \\
\midrule
Question diversity (tokens) & 11,250 \\
Answer diversity (tokens) & 14,890 \\
\midrule
Input length distribution (tokens) &  \\
\midrule
{[}959, 3,717{]} & 49.98\% \\
{[}3,717, 13,505{]} & 40.01\% \\
{[}13,505, 14,124{]} & 4.94\% \\
{[}14,124, 135,055{]} & 5.07\% \\
\midrule
Median / mean tokens & 4,654 / 4,963 \\
\bottomrule
\end{tabular}
\end{table}

\subsection{Further Dataset Statistics}
\label{sec:Further_Dataset_Statistics}
Tab.~\ref{tab:dataset_stats} summarizes the statistics of the multi-turn open-ended subset of \benchmarkname.
The dataset comprises 6,169 multi-turn conversations grounded on 2,427 unique images, with an average of 4.25 turns per session, reflecting realistic multi-round interaction patterns.
Questions and answers exhibit substantial lexical diversity, as measured by MTLD, indicating rich linguistic variability and reduced template bias.
In addition, the input length distribution spans a wide range of token counts, with a median of 4,654 tokens and a maximum exceeding 135k tokens, highlighting the presence of long-context scenarios that challenge models’ memory, context tracking, and cross-turn reasoning capabilities.

\begin{figure*}[!t]
    \centering
    \includegraphics[width=\linewidth]{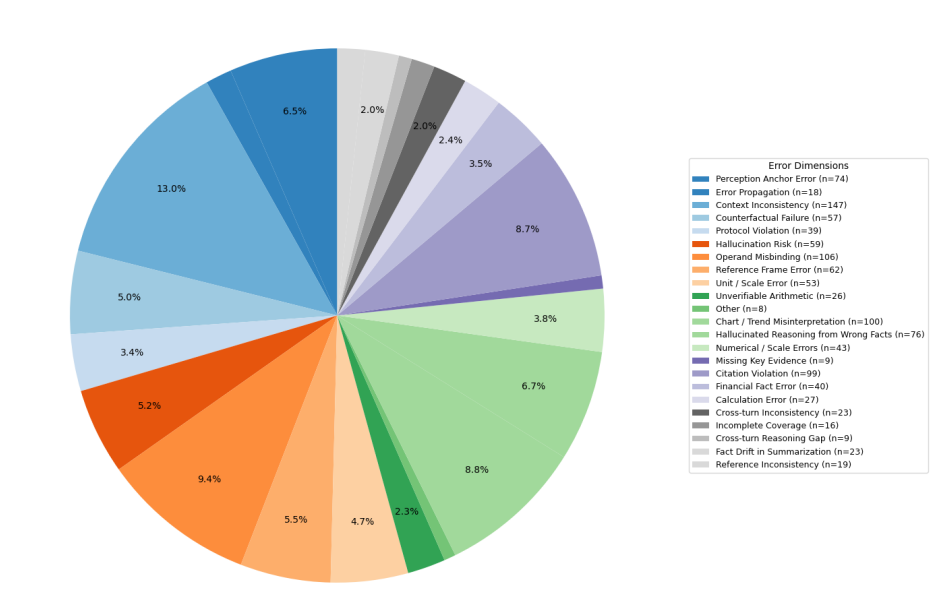}
    \caption{This figure presents the distribution of error types, based on 1133 responses sampled from Gemini 3 Pro. Each response was manually reviewed, and categorized into distinct error types.}
    \label{fig:error}
\end{figure*}

\begin{table*}[htbp]
\centering
\setlength{\tabcolsep}{5pt}
\begin{tabular}{lcccccc}
\toprule
\textbf{Model} & \textbf{Tool} & \textbf{Reason} & \textbf{Total} & \textbf{Recall (\%)} & \textbf{F1 (\%)} & \textbf{EMR} \\
\midrule
Gemini\_3\_Flash\_Preview & \textbf{11.2653} & \textbf{21.7959} & \textbf{62.9000} & \textbf{44.9245} & \textbf{42.2057} & \textbf{0.1020} \\
ChatGPT-5                     & 8.0816 & 16.2857 & 44.7755 & 39.7694 & 30.3771 & 0.0302 \\
Grok                      & 9.3061 & 12.3469 & 36.9592 & 36.0402 & 34.8976 & 0.0408 \\
Gemini                    & 4.8333 & 20.3095 & 54.9048 & 25.1969 & 19.2095 & 0.0331 \\
ChatGPT-3                        & 6.6327 & 13.3469 & 35.2857 & 22.7796 & 22.4422 & 0.0204 \\
\midrule
Qwen3-VL-235B-Thinking    & 7.4231 & 15.1245 & 41.5122 & 36.8541 & 33.9102 & 0.0321 \\
GLM-4.5V-108B             & 5.8541 & 11.9210 & 32.4353 & 28.1245 & 25.3210 & 0.0275 \\
Qwen3-VL-32B-Instruct     & 4.5124 & 9.1541  & 25.1124 & 22.3120 & 19.7852 & 0.0142 \\
InternVL3-78B             & 4.1245 & 8.3541  & 22.8857 & 20.2415 & 17.9214 & 0.0118 \\
InternVL2.5-40B           & 3.0142 & 6.1425  & 16.8346 & 15.1241 & 13.2145 & 0.0075 \\
\midrule
\bottomrule
\end{tabular}
\caption{Model-wise evaluation statistics on financial agent task.}
\label{tab:model_scores}
\end{table*}

\subsection{Dataset Examples}
\label{sec:finmtm_examples}

FinMTM is designed to evaluate financial VLMs under heterogeneous supervision signals and interaction patterns, spanning objective selection, multi-turn report-grounded dialogue, and tool-augmented agentic planning.
Fig.~\ref{fig:case1} illustrates the \emph{objective} setting, including single-choice and multi-choice questions grounded in technical-indicator charts.
Such problems emphasize fine-grained visual perception and option-level discrimination, serving as a controlled testbed for factual grounding and local reasoning.

Beyond objective evaluation, FinMTM includes \emph{multi-turn open-ended} report-centric dialogues (Fig.~\ref{fig:case3}), where the model must extract numbers from multi-page reports, cite evidence across pages, perform intermediate computations, and verify statements over successive turns.
This setting stresses long-context stability, cross-page evidence binding, and arithmetic reliability under explicit citation requirements.

Finally, Fig.~\ref{fig:case2} showcases a \emph{multi-turn planning-agent} session, where the model must first interpret the sector trend from charts, then plan and execute tool calls to retrieve candidate entities under business/market constraints, and ultimately conduct consistency checking against company-level financial indicators.
This agentic setting highlights long-horizon planning, tool-use orchestration, and entity disambiguation; these capabilities are critical to real-world financial workflows but remain challenging for current VLMs.

\section{Error Analysis}
\label{Error Analysis supp}
Although Gemini shows strong overall performance on \benchmarkname, we conduct targeted case studies on low-scoring examples to analyze representative failure cases. As shown in Fig.~\ref{fig:error}, Gemini exhibits systematic errors such as chart misinterpretation and cross-turn inconsistency.

\subsection{\texorpdfstring{Comprehension ($L_1$)}{Comprehension (L1)}}

An analysis of low-score cases reveals that the dominant source of failure lies in inaccurate {financial visual perception}. In early turns, the model frequently misidentifies critical perceptual anchors in financial charts, such as global extrema, their corresponding timestamps, or precise value locations. These incorrect perceptual outputs are subsequently treated as fixed financial facts and propagated into later multi-step reasoning.
Due to the absence of explicit mechanisms for re-verifying or correcting perceptual results against the visual input, initial perception errors persist across turns and are further amplified during downstream operations such as difference estimation and counterfactual adjustment. In addition, when inconsistencies arise between user-provided descriptions and visual evidence, the model often fails to reconcile them, passively accepting erroneous contextual cues and thereby exacerbating perceptual drift.

Overall, low-score samples highlight fundamental limitations in the model’s ability to maintain {stable and verifiable financial perceptual states}, which critically undermines its performance in perception-grounded multi-step reasoning over financial charts.

\subsection{\texorpdfstring{Calculation ($L_2$)}{Calculation (L2)}}

An analysis of low-scoring cases  reveals that failures are not primarily caused by deficient reasoning structures, but rather by {incorrect binding of numerical values and reference objects during logical computation}. In most cases, the model produces a coherent reasoning chain, yet selects incorrect operands prior to computation, such as confusing different curves or subplots, or mistaking local extrema for global ones, which renders subsequent difference or ratio calculations invalid.

Moreover, when handling computations involving peak--trough differences or drawdowns, the model often fails to explicitly establish a global reference frame and directly performs arithmetic based on incorrect or incomplete baselines. Errors in unit and scale handling (\textit{e.g.}, inconsistent conversions between percentages and basis points) further amplify numerical deviations. In some cases, the model reports precise numerical results despite the absence of explicit axis scales or value annotations, leading to unverifiable calculations.

Overall, these low-scoring cases indicate that the model’s limitations lie in insufficient modeling of {pre-computational constraints}, including variable binding, reference validation, and unit consistency, resulting in a characteristic failure mode of logically coherent yet numerically inaccurate reasoning.

\subsection{\texorpdfstring{Self-correction ($L_3$)}{Self-correction (L3)}}

We analyze low-score cases in the $L_3$ multi-turn visual reasoning task to identify common failure patterns. The dominant issue is {misinterpretation of visual evidence}, including incorrect trend direction, legend association, or key inflection points. As visual understanding forms the factual basis of subsequent reasoning, such errors cause the entire analysis to rely on incorrect assumptions.
Moreover, the model exhibits deficiencies in {cross-turn consistency}, as its predictions across successive turns may rely on mutually inconsistent assumptions. It also shows limited \textbf{self-correction ability}, often failing to update or retract earlier assumptions even when later turns provide explicit corrective cues. Overall, low scores are primarily driven by deficiencies in visual grounding and error correction rather than limitations in linguistic fluency or reasoning structure.

\begin{figure*}[!t]
    \centering
    \includegraphics[width=\linewidth]{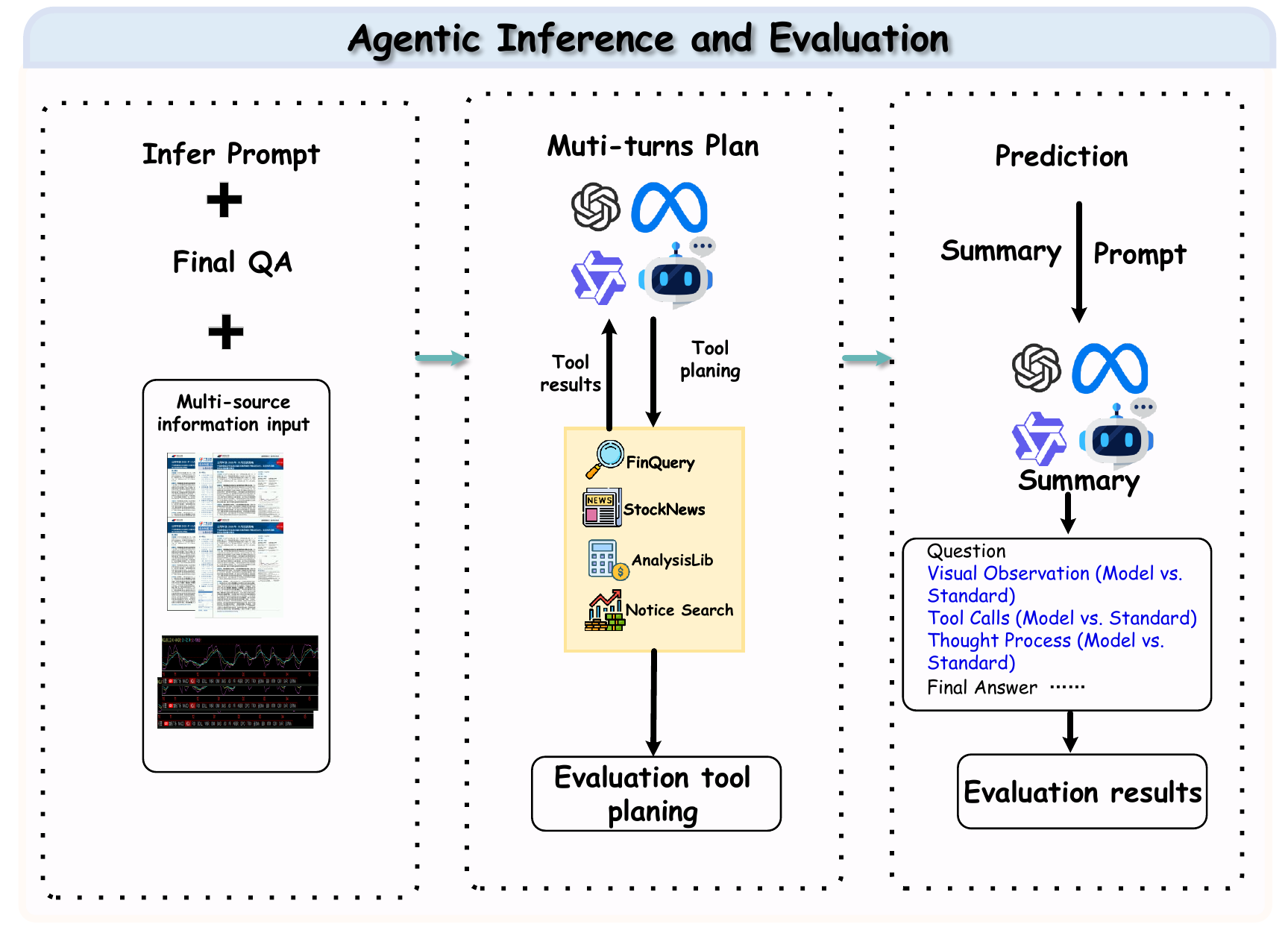}
    \caption{Illustration of the agentic evaluation pipeline.}
    \label{fig:agentic_eval}
\end{figure*}

\begin{figure*}[!t]
    \centering
    \includegraphics[width=\linewidth]{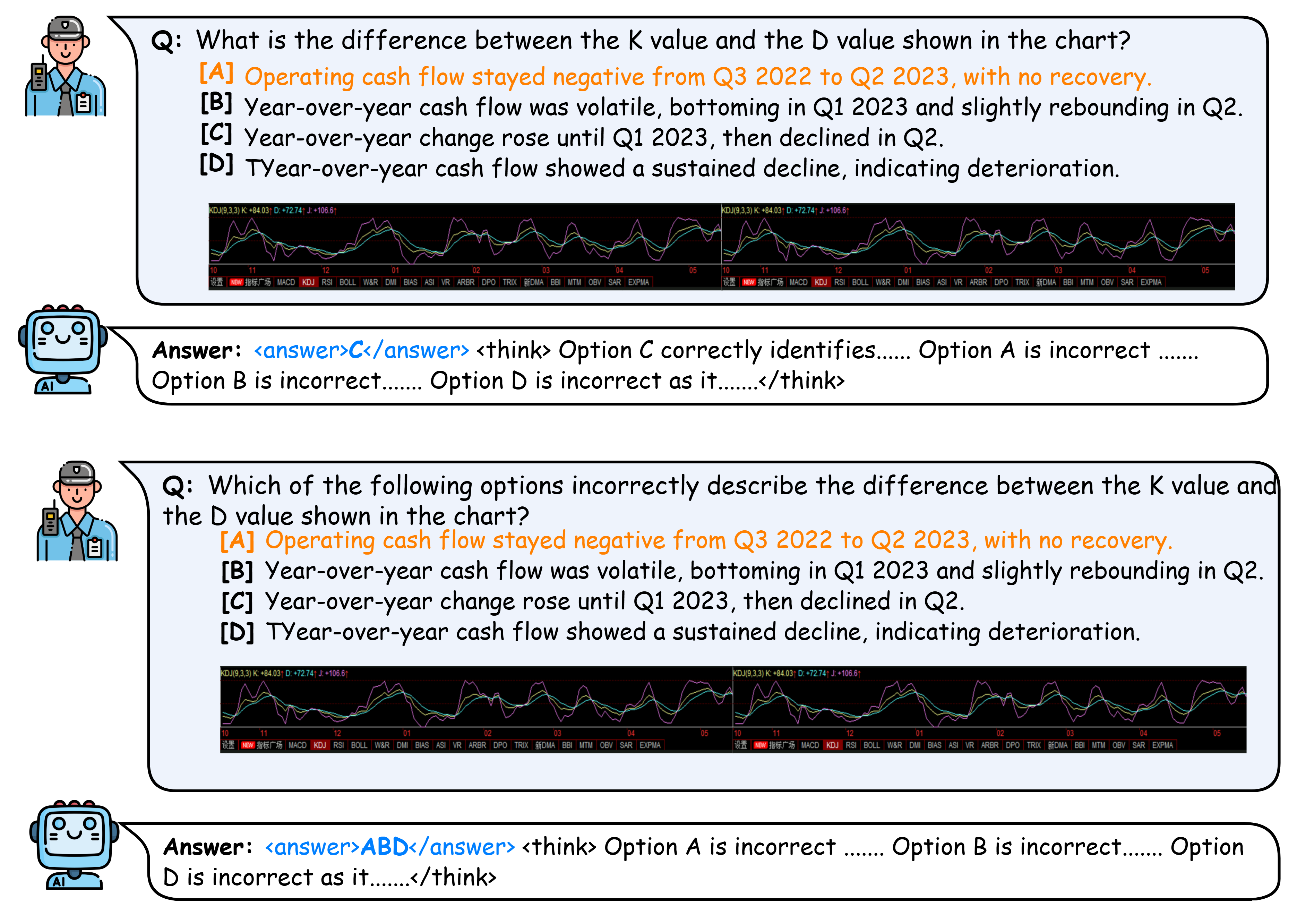}
    \caption{Representative FinMTM objective questions: single-choice (top) and multi-choice (bottom) on technical-indicator interpretation (K--D curves) grounded in financial charts.}

    \label{fig:case1}
\end{figure*}
\begin{figure*}[!t]
    \centering
    \includegraphics[width=\linewidth]{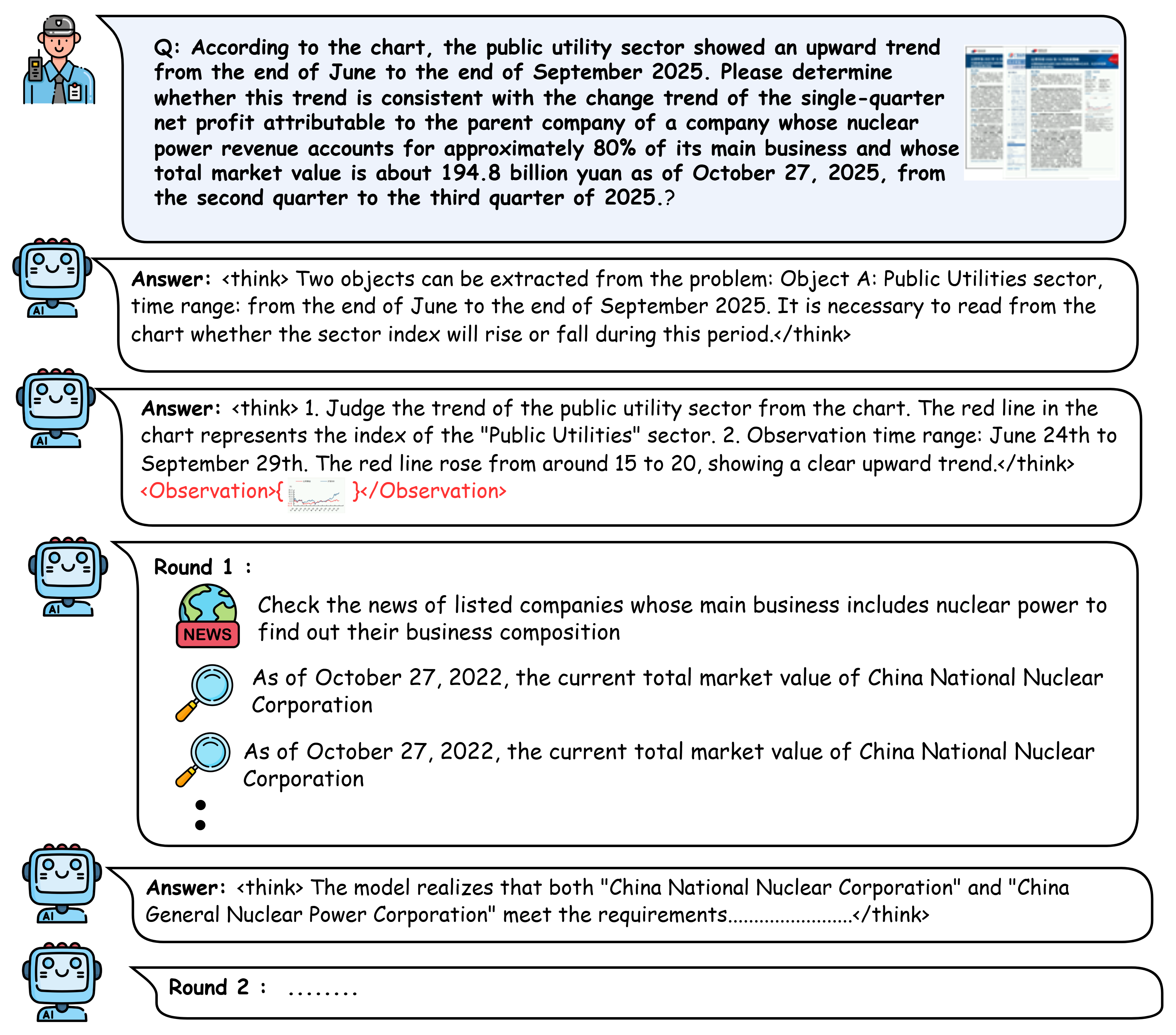}
    \caption{A FinMTM multi-turn planning-agent case illustrating chart-grounded reasoning, tool-assisted entity search, and consistency verification across heterogeneous financial evidence.}
    \label{fig:case2}
\end{figure*}
\begin{figure*}[!t]
    \centering
    \includegraphics[width=\linewidth]{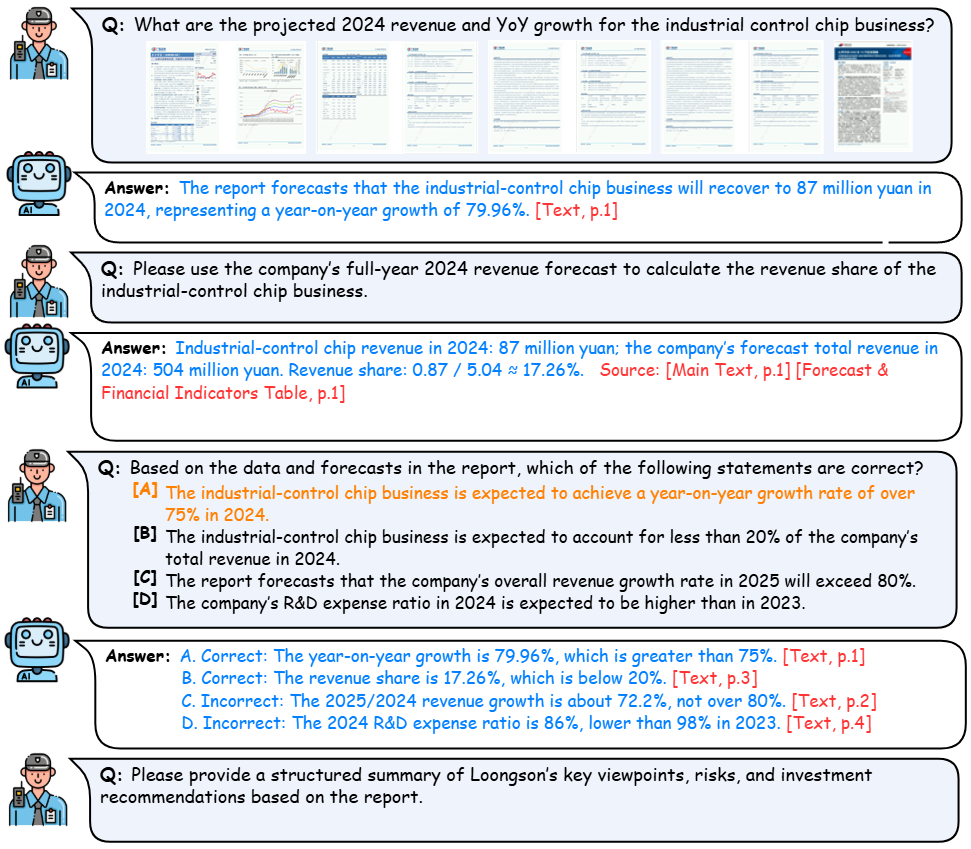}
    \caption{A representative FinMTM multi-turn open-ended dialogue grounded in multi-page financial reports, requiring cross-page evidence citation, numerical extraction, intermediate calculation, and statement verification across turns.}

    \label{fig:case3}
\end{figure*}
\subsection{\texorpdfstring{Memory ($L_4$)}{Memory (L4)}}
An inspection of low-scoring samples in the $L_4$ setting reveals that most failures are not caused by deficiencies in visual perception or logical reasoning, but rather by violations of strict evaluation constraints and financial expression errors. The dominant source of failure stems from non-compliance with the citation-reliance rule, which requires explicit and structured references (\textit{e.g.}, \texttt{[Source, Page X]}) at every dialogue turn. Missing or incomplete citations invalidate the entire multi-turn trajectory, even when individual responses are factually correct.

In addition to citation-related issues, a subset of errors arises from financial inaccuracies, including profit--loss direction confusion, unit misinterpretation, and incorrect growth-rate or percentage calculations. These errors are amplified in multi-turn scenarios, where early numerical deviations propagate through subsequent reasoning steps. Furthermore, some failures are attributed to cross-turn inconsistencies, where later summarization omits or contradicts previously established facts.

Overall, the observed low-score cases primarily reflect challenges in adhering to strict verification-oriented protocols and maintaining financial rigor across turns, rather than fundamental limitations in multimodal understanding or reasoning capability.

\section{Further Analysis on Financial Agent}
\label{sec:Further Analysis agent}

Fig.~\ref{fig:agentic_eval} shows the inference and evaluation pipeline of financial agent task. In the agent-oriented experiments on FinMTM, we observe that the introduction of the \textbf{}{fuzz} strategy consistently degrades performance across all evaluated models, highlighting the increased difficulty of decision-making under reduced semantic priors. 
A detailed examination of failure cases reveals several key observations.

\noindent\textbf{Incompatibility with Agent Frameworks Under Fuzzed Inputs.}
For a subset of queries, agents fail to return valid outputs when operating under the fuzz setting.
Stronger models, such as Gemini, are often able to recover by reformulating queries or adjusting interaction strategies, whereas weaker models are more prone to entering irreversible failure states.

\noindent\textbf{Limited Agentic Capability of Small-scale Models.}
Smaller models frequently exhibit redundant or ineffective tool usage, repeatedly invoking the same tool or querying identical information through multiple tools.
Such behavior indicates deficiencies in planning, memory, and internal state management within the agent loop.

\noindent\textbf{Amplified Agent Failures Caused by Weak Visual Perception.}
Insufficient fine-grained visual understanding further exacerbates agent-level failures.
In many cases, small models attempt to retrieve information via tools that could have been directly inferred from visual inputs (\textit{e.g.}, trend patterns in industry charts).
Since such information is unavailable through tool calls, these agents often trigger the forced summarization mechanism and terminate prematurely.

\noindent\textbf{Superior Financial Agent Performance of Gemini 3 Flash.}
Among proprietary models, Gemini 3 Flash demonstrates notably stronger financial agent capabilities.
However, in pure reasoning settings without agent interactions, closed-source models show much smaller performance gaps, suggesting their advantage mainly comes from stronger agent execution rather than raw reasoning.

As summarized in Tab.~\ref{tab:model_scores}, models exhibit clear capability stratification in the FinMTM financial agent scenario.
Overall, closed-source models significantly outperform open-source counterparts in terms of comprehensive agent performance.
In particular, Gemini 3 Flash-Preview leads across core metrics such as \emph{Tool}, \emph{Reason}, and \emph{Total}, indicating strong robustness in tool invocation stability, reasoning quality, and result alignment.
While Gemini and ChatGPT-5 demonstrate strong reasoning and information coverage, they still suffer from noticeable performance degradation in execution-alignment metrics (\textit{e.g.}, Recall, F1, and EMR), reflecting limitations in translating reasoning ability into effective agent execution.
Among open-source models, Qwen3-VL-235B-Thinking stands out, achieving performance comparable to some closed-source systems and demonstrating relatively strong agent execution and alignment capabilities.
In contrast, smaller-scale models (\textit{e.g.}, Qwen3-VL-32B and the InternVL series) consistently underperform on both \emph{Tool} and \emph{Reason} metrics, resulting in substantially limited overall performance.
These results indicate that, in complex financial agent scenarios, model scale and agent-level planning and state management capabilities remain key determinants of performance.

\begin{figure*}[p]
\centering
\begin{tcolorbox}[
    title={Inference Prompt for Objective Question},
    colback=gray!6,
    colframe=black!40,
    boxrule=0.6pt,
    fonttitle=\bfseries,
    left=6pt,
    right=6pt,
    top=4pt,
    bottom=4pt
]

\small
\textbf{System Instruction.}
You are a financial expert. Read the question and the image.
Output ONLY a JSON object, with NO extra text, NO explanations, and NO backticks.

\medskip
\textbf{Fields required:}
\begin{itemize}
    \item \texttt{answer}: the final choice from the options
    \item \texttt{reason}: briefly explain your reasoning
\end{itemize}

\medskip
\textbf{Question.}
How many times do the two curves cross after May 1st, 2024 in the MACD chart?
\medskip

\textbf{Image.}
{Url}

\medskip
\textbf{Options.}
\begin{enumerate}[label=\Alph*.]
    \item Four crossings after 24.05
    \item Two crossings after 24.05
    \item Three crossings after 24.05
    \item One crossing after 24.05
\end{enumerate}

\medskip
\textbf{Output Format.}
\begin{lstlisting}[basicstyle=\ttfamily\footnotesize]
{"answer":"C","reason":"The white and yellow lines cross three times."}
\end{lstlisting}

\end{tcolorbox}
\caption{An example inference prompt for objective questions on financial chart.}
\label{box:QA-prompt}
\end{figure*}

\begin{figure*}[p]
\centering
\begin{tcolorbox}[
    title={Multi-turn Inference Prompt for Open-ended Question},
    colback=gray!6,
    colframe=black!40,
    boxrule=0.6pt,
    fonttitle=\bfseries,
    left=6pt,
    right=6pt,
    top=4pt,
    bottom=4pt
]
\small

\textbf{System Instruction.}
You are a financial expert.
Read the question, the image, and the conversation history.
For each turn, output ONLY the answer to the current question.
Do NOT include explanations unless explicitly required.

\medskip
\textbf{Image.}
\texttt{url}

\medskip
\textbf{Conversation History and Current Turn.}

\medskip
\textbf{Turn T1}
\begin{itemize}[leftmargin=1.2em]
    \item \textbf{Question:}
    Locate the global maximum of the red polyline in the chart,
    denote it as point A1, and report its approximate date.
\end{itemize}

\medskip
\textbf{Turn T2}
\begin{itemize}[leftmargin=1.2em]
    \item \textbf{Context:}
    Point A1 is the global maximum of the red polyline,
    with an approximate date of 2023-12-22.
    \item \textbf{Question:}
    At point A1, how much higher is the value of the red polyline
    compared to the yellow polyline?
\end{itemize}

\medskip
\textbf{Turn T3}
\begin{itemize}[leftmargin=1.2em]
    \item \textbf{Context:}
    Point A1 is the global maximum of the red polyline,
    dated approximately 2023-12-22,
    and the original red--yellow difference is about 60.0.
    \item \textbf{Question:}
    Assume that due to data correction, the actual value of the red
    polyline at point A1 is 10\% lower than the observed value.
    Recalculate the difference between the red and yellow polylines.
\end{itemize}

\medskip
\textbf{Turn T4}
\begin{itemize}[leftmargin=1.2em]
    \item \textbf{Context:}
    Point A1 is the global maximum of the red polyline,
    dated approximately 2023-12-22,
    and the corrected red--yellow difference is about 56.0.
    \item \textbf{Question:}
    At the same timestamp as point A1,
    is the value of the green polyline below $-20.0$?
\end{itemize}

\medskip
\textbf{Expected Output (per turn).}
\begin{lstlisting}[basicstyle=\ttfamily\footnotesize]
T1: 2023-12-22
T2: 60.0
T3: 56.0
T4: Yes
\end{lstlisting}

\end{tcolorbox}
\caption{An example zero-shot inference prompt for open-ended questions.}
\label{box:Multiturn-prompt}
\end{figure*}

\begin{figure*}

\begin{tcolorbox}[
  title={Multi-turn Financial Agent Inference Prompt},
  colback=gray!6,
  colframe=black!40,
  boxrule=0.6pt,
  fonttitle=\bfseries,
  left=6pt, right=6pt, top=4pt, bottom=4pt
]
\small

\textbf{(1) System Instruction (Shared Across All Rounds).}
The agent is instructed to act as a \emph{financial multi-round analysis agent}.
Its goal is to iteratively plan MCP tool calls and integrate tool outputs
with visual evidence extracted from the input image to answer the query.

\medskip
\textbf{Available Tools.}
\begin{itemize}[leftmargin=1.2em]
  \item \textbf{FinQuery}: financial indicators, prices, and fundamentals.
  \item \textbf{Search}: general entity or concept lookup.
  \item \textbf{StockNews}: financial news retrieval.
  \item \textbf{ReportQuery}: analyst report inspection.
\end{itemize}

\medskip
\textbf{Output Contract.}
At each planning step, the agent must output \emph{only one JSON object}:
\begin{lstlisting}[basicstyle=\ttfamily\footnotesize]
{
  "Thought": "high-level reasoning and planning",
  "VisualObservation": ["evidence read from the image"],
  "ActionTrace": [
    {"tool": "FinQuery", "query": "..."},
    {"tool": "Search", "query": "..."}
  ]
}
\end{lstlisting}
If the task is completed, the agent must emit \texttt{<FINISHED>}
and provide the final conclusion.

\medskip
\hrule
\vspace{4pt}

\textbf{(2) First-round Prompt.}
In the first round, the agent receives only:
(i) the system instruction,
(ii) the input question, and
(iii) the image (if available).
No prior reasoning or tool feedback is provided.

\medskip
\hrule
\vspace{4pt}

\textbf{(3) Iterative Prompt for Subsequent Rounds.}
For round $t>1$, the prompt is augmented with evidence accumulated
from previous rounds:

\begin{lstlisting}[basicstyle=\ttfamily\footnotesize]
<System Instruction>
 
Previous Thought:
<Thought_{t-1}>

Visual Observation:
<VisualObservation>

Tool Feedback:
<ToolFeedback>

Continue planning based on the above information.
Do NOT repeat tool calls.
If the information is sufficient, output <FINISHED> and conclude.
\end{lstlisting}

The same image is re-attached at every round to compensate for the
stateless image-to-text interface.

\medskip
\hrule
\vspace{4pt}

\textbf{(4) Forced Summarization Prompt (Termination Fallback).}
If the agent fails to emit \texttt{<FINISHED>} within the maximum number
of planning rounds, a forced summarization prompt is triggered.
In this stage, \emph{no further tool calls are allowed}.

\begin{lstlisting}[basicstyle=\ttfamily\footnotesize]
Based on the following visual observations and tool feedback,
summarize the final answer.

Question: <query>

Visual Observation: <VisualObservation>

Tool Feedback: <ToolFeedback>

Provide the final conclusion and output <FINISHED>.
\end{lstlisting}

This prompt explicitly tests the agent’s ability to synthesize
multi-round evidence into a coherent final decision under a strict
termination constraint.

\end{tcolorbox}
\caption{An example zero-shot inference prompt for multi-turn financial agent.}
\label{box:agent-prompt}
\end{figure*}

\end{document}

%% file: table/compare.tex
\begin{tabular}{lcccccccc}
\toprule[1.2pt]

\rowcolor{gray!15}
 & \multicolumn{2}{c}{\textbf{Objective Questions}}
 & \multicolumn{4}{c}{\textbf{Open-Ended Ques.}}
 & \multicolumn{2}{c}{\textbf{Financial Agent}} \\ 

\cmidrule(lr){2-3}
\cmidrule(lr){4-7}
\cmidrule(lr){8-9}

\rowcolor{gray!15}
\multirow{-2.5}{*}{\textbf{Method}} 
 & \textbf{Single.} 
 & \textbf{Multi.} 
 & \textbf{Com.} 
 & \textbf{Cal.} 
 & \textbf{SelfCorr.} 
 & \textbf{Mem.} 
 & \textbf{w fuzz}      
 & \textbf{w/o fuzz} \\  

\midrule[1pt]

\multicolumn{9}{l}{\textit{Proprietary Models}} \\
\midrule[0.5pt]
ChatGPT-4o~\cite{gpt4o}     
 & 79.3 & 49.1 & 77.2 & 76.8 & 46.2 & 38.9 & 29.7 & 34.8 \\

ChatGPT-o3*~\cite{openai2025o3systemcard}    
 & 85.8 & 73.3 & 83.8 & 78.6 & 52.8 & 43.6 & 31.4 & 35.2  \\

ChatGPT-5*~\cite{openai2025gpt5systempdf}     
 & 89.0 & \textbf{79.6} & \underline{86.9} & \underline{80.7} & \underline{56.9} & \underline{46.7} & 35.9 & 49.7 \\

Gemini 3 Flash~\cite{google2025gemini3flashsystemcard}     
 & \underline{91.9} & 78.1 & 82.2 & 76.0 & 55.4 & 41.6 & \textbf{53.6} & \textbf{62.6}  \\

Grok-4-fast-non-reasoning*~\cite{oracle_xai_grok4fast}    
 & 71.0 & 46.8 & 66.0 & 61.2 & 39.9 & 24.8 & 30.2 & 39.7  \\

Gemini 3 Pro~\cite{google2025gemini3prosystemcard}    
 & \textbf{92.1} & \underline{78.4} & \textbf{87.5} & \textbf{82.8} & \textbf{58.8} & \textbf{48.5} & \underline{48.3} & \underline{54.3} \\

\midrule[1pt]

\multicolumn{9}{l}{\textit{InternVL Series}} \\
\midrule[0.5pt]
InternVL2.5-8B~\cite{internvl25}   
 & 63.8 & 25.7 & 55.1 & 49.2 & 26.5 & 16.7 & 8.4  & 10.5 \\

InternVL2.5-26B~\cite{internvl25}   
 & 70.5 & 31.3 & 61.7 & 57.7 & 32.3 & 22.8 & 11.2 & 14.0 \\

InternVL2.5-40B~\cite{internvl25}   
 & 72.3 & 35.2 & 66.1 & 64.6 & 36.2 & 26.7 & 13.5 & 16.8 \\

InternVL3-78B~\cite{internvl3}     
 & 75.6 & 42.4 & 76.2 & 77.6 & 43.6 & 32.6 & 18.2 & 22.8 \\

\midrule[0.5pt]
\multicolumn{9}{l}{\textit{Other VL Series}} \\
\midrule[0.5pt]
MiMo-VL-7B~\cite{xiaomi2025mimo}       
 & 61.1 & 21.4 & 75.1 & 75.4 & 47.2 & 39.9 & 20.2 & 25.5 \\

GLM4.5V-108B~\cite{hong2025glm45v}     
 & 73.7 & 51.0 & 85.4 & 79.6 & 51.1 & 42.2 & 26.5 & 32.4 \\

\midrule[0.5pt]
\multicolumn{9}{l}{\textit{Qwen VL Series}} \\
\midrule[0.5pt]
Qwen2.5-VL-3B~\cite{qwen25vl}       
 & 64.5 & 16.4 & 68.2 & 67.7 & 40.5 & 27.6 & 9.4  & 11.9 \\

Qwen2.5-VL-7B~\cite{qwen25vl}       
 & 73.4 & 24.1 & 74.3 & 73.4 & 43.1 & 33.9 & 11.1 & 14.2 \\

\midrule[0.3pt]
Qwen3-VL-4B-Instruct~\cite{qwenvl3}   
 & 73.3 & 34.2 & 74.5 & 71.2 & 39.5 & 25.9 & 15.1 & 19.1 \\

Qwen3-VL-4B-Thinking~\cite{qwenvl3}   
 & 66.1 & 24.3 & 71.2 & 68.5 & 42.5 & 31.0 & 12.8 & 15.6 \\

\midrule[0.3pt]
Qwen3-VL-30B-A3B-Instruct~\cite{qwenvl3} 
 & 77.2 & 47.3 & 82.1 & 76.5 & 42.5 & 33.7 & 16.2 & 20.8 \\ 

Qwen3-VL-30B-A3B-Thinking~\cite{qwenvl3} 
 & 71.5 & 49.4 & 80.7 & 67.1 & 44.2 & 35.1 & 18.9 & 23.3 \\

\midrule[0.3pt]
Qwen3-VL-32B-Instruct~\cite{qwenvl3}   
 & 84.5 & 39.9 & 84.3 & 80.7 & 50.8 & 40.3 & 19.6 & 25.1 \\ 

Qwen3-VL-32B-Thinking~\cite{qwenvl3}   
 & 83.4 & 46.5 & 80.3 & 68.6 & 43.5 & 33.7 & 23.2 & 28.6 \\

\midrule[0.3pt]
Qwen3-VL-235B-A22B-Instruct~\cite{qwenvl3}  
 & 81.3 & 48.5 & 85.5 & 80.9 & 54.5 & 41.5 & 32.1 & 38.7 \\

Qwen3-VL-235B-A22B-Thinking~\cite{qwenvl3}  
 & 80.5 & 42.3 & 84.5 & 79.4 & 52.5 & 43.0 & 35.2 & 41.5 \\

\bottomrule[1.2pt]
\end{tabular}

%% file: table/compare_v2.tex
\begin{tabular}{lcccccc}
\toprule
\textbf{Model} & \textbf{VP} & \textbf{FL} & \textbf{DA} & \textbf{CMV} & \textbf{TA} & $\mathbf{S_e}$ \\
\midrule
Qwen3-VL-4B-Instruct              
& 68.60 & 49.80 & 35.40 & 45.50 & 72.10 & 25.97 \\
Qwen2.5-7B                      
& 78.00 & 59.90 & 45.60 & 59.60 & 83.50 & 26.53 \\

Qwen3-VL-30B-A3B-Instruct         
& 76.00 & 59.50 & 44.40 & 56.60 & 82.40 & 35.32 \\
Qwen3-VL-30B-A3B-Thinking       
& 69.20 & 54.10 & 40.40 & 51.50 & 75.00 & 33.41 \\
Qwen3-VL-32B-Instruct              
& 84.30 & 73.70 & 59.50 & 71.10 & 92.00 & 42.86 \\
Qwen3-VL-32B-Thinking              
& 76.50 & 59.40 & 44.10 & 56.30 & 82.40 & 40.12 \\
Qwen3-VL-235B-A32B-Instruct        
& 87.50 & 78.00 & 62.90 & 73.80 & 92.80 & 52.41 \\
Qwen3-VL-235B-A32B-Thinking        
& 85.80 & 77.20 & 61.40 & 74.50 & 92.10 & 50.57 \\
\bottomrule
\end{tabular}